\documentstyle[a4,11pt]{article}
\title{A Note on the Complexity of Restricted Attribute-Value
Grammars}
\author{Leen Torenvliet%
\thanks{The author was supported in part by HC\&M grant
	ERB4050PL93-0516.}
\and Marten Trautwein%
\thanks{The author was supported by the Foundation for language,
	speech and logic (TSL), which is funded by the Netherlands
	organization for scientific research (NWO)}}
\date{\small\em University of Amsterdam\\Department of Mathematics
and Computer Science\\Plantage Muidergracht 24\\1018 TV Amsterdam}
\newcommand{\C}[1]{\mathord{\it #1}}

\newenvironment{deflike}[2]{\refstepcounter{#2}
\begin{list}{}
{\setlength{\topsep}{1em}
\setlength{\itemindent}{\leftmargin}
\addtolength{\itemindent}{-.5em}
\setlength{\leftmargin}{1em}
\setlength{\leftmargin}{1em}
\setlength{\rightmargin}{1em}}
\item[{\bf #1 \csname the#2\endcsname}:]}{\end{list}}
\newenvironment{define}{\begin{deflike}{Definition}{defctr}}
{\end{deflike}}
\newcommand{\prs}{\par}
\newenvironment{proof}{\noindent{\it Proof}.}{\hfill\mbox{$\Box$}
\prs\vspace{3mm}\prs}
\newcommand{\lng}[1]{\mathopen|#1\mathclose|}
\newtheorem{thm}{Theorem}[section]
\newtheorem{cor}[thm]{Corollary}
\newcommand{\Cat}{\mathord{\rm Cat}}
\newcommand{\Lex}{\mathord{\rm Lex}}
\newtheorem{lem}[thm]{Lemma}

\newcommand{\fn}[1]{\mbox{\sc #1}}
\newcommand{\fv}[1]{\mbox{\it #1}}
\newcommand{\avm}[1]{\left[\begin{array}{ll} #1 \end{array}\right]}
\newcommand{\avg}[2]{\begin{array}[#1]{l@{}l} #2 \end{array}}

\begin{document}
\maketitle

\begin{abstract}
The recognition problem for attribute-value grammars(AVGs) was shown
to be undecidable by Johnson in 1988. Therefore, the general
form of AVGs is of no practical use. In this paper we study
a very restricted form of AVG, for which the recognition problem
is decidable (though still $\C{NP}$-complete), the R-AVG. We show that
the R-AVG formalism captures all of the context free languages and
more, and introduce a variation on the so-called {\em off-line
parsability constraint\/}, the {\em honest
parsability constraint\/}, which lets different types of
R-AVG coincide precisely with well-known time complexity classes.
\end{abstract}
\section{Introduction}
Although a universal feature theory does not exist, there is
a general understanding of its objects. The objects of feature
theories are abstract linguistic objects, e.g., an object ``sentence,''
an object ``masculine third person singular,'' an object ``verb,'' an
object ``noun phrase.'' These abstract objects have properties like
``tense,'' ``number,'' ``predicate,'' ``subject.''
The values of these properties
are either atomic, like ``present'' and ``singular,'' or abstract
objects, like ``verb'' and ``noun-phrase.''
The abstract objects are fully described by their properties and
their values. Multiple descriptions for the properties and values
of the abstract linguistic objects are presented in the literature.
Examples are:

\begin{enumerate}
\item Feature graphs, which are labeled rooted directed acyclic graphs
$G=(V,A)$, where
$F$ is a collection of labels,
a sink
in the graph represents an atomic value and the labeling function
is an injective function $f:V\times A\mapsto F$.
\item Attribute-value matrices,
which are matrices in which the
entries consist of an attribute and a value or
a reentrance symbol. The values are either atomic or
attribute-value matrices.
\end{enumerate}

{}From a computational point of view, all descriptions that are used in
practical problems are equivalent. Though there exist some theories
with a considerably higher expressive power \cite{blackburn.ea:93}.
For this paper we adopt the feature graph description, which we will
define somewhat more formal in the next section.
Attribute Value Languages(AVL)~\cite{smolka:92} consist of
sets of logical formulas that describe classes of feature graphs,
by expressing constraints on the type of paths that can exist within
the graphs. To wit: In a sentence like ``a man walks'' the edges
labeled with ``person'' that leave the nodes
labeled ``a man'' and ``walks'' should both
end in a node labeled ``singular.'' Such a constraint is called
a ``path equation'' in the attribute-value language.

A rewrite grammar~\cite{chomsky:56} can be enriched with an
AVL to construct an Attribute Value
Grammar(AVG), which consists of pairs of rewrite-rules and logical
formulas. The rewrite rule is applicable to a production (nonterminal)
only if the logical formula that expresses the relation between
left- and right-hand side of the rule evaluates to true.
The recognition problem for attribute-value grammars can be
stated as: Given a grammar $G$ and a string $w$ does there
exist a derivation in $G$, that respects the constraints
given by its AVL, and that ends in $w$.
As the intermediate productions correspond
to feature graphs this question can also be formulated as
a question about the existence of a consistent sequence of
feature graphs that results in a feature graph describing $w$.
For the rewrite grammar, any formalism in the Chomsky hierarchy
(from regular to type $0$) can be chosen. From a computational point
of view it is of course most desirable to restrict oneself to
a formalism that on the one hand gives enough expressibility
to describe a large fragment of the (natural) language, and
on the other hand is restrictive enough to preserve feasibility.
For a discussion on the linguistic significance of such restrictions,
see \cite{perrault:84}.

Johnson~\cite{johnson:88} proved that attribute-value grammars
that are as restrictive as being equipped with a rewrite grammar
that is regular can already give rise to an undecidable recognition
problem.
Obviously, to be of any practical use, the
rewrite grammar or the attribute-value language must be more restrictive.
Johnson proposed to add the {\em off-line parsability constraint\/},
which is respected if the rewrite grammar has no chain- or
$\epsilon$-rules. Then, the number of applications in a production
is linear and the size of the structure corresponding to the
partial productions is polynomial. Hence as by a modification of
Smolka's algorithm~\cite{smolka:92} consistency of intermediate
steps can be checked in quadratic time, the complexity of
the recognition problem can at most be (nondeterministic) polynomial
time. This observation was made in~\cite{lp-95-01}, which
also has an $\C{NP}$-hardness proof of the recognition problem.

We further investigate the properties of these restricted AVGs (R-AVGs).
In the next section, we give some more formal definitions and notations.
In Section~\ref{WGC} we show that the class
of languages generated by an R-AVG (R-AVGL) includes
the class of context free languages (CFL). It follows that any easily
parsable class of languages (like CFL) is a proper
subset of R-AVGL, unless $\C{P}=\C{NP}$.
Likewise, R-AVGL is a proper subset of the class of context
sensitive languages, unless $\C{NP}=\C{PSPACE}$.
In Section~\ref{HPC}
we propose a further refinement on the off-line parsability constraint,
which allows R-AVGs that respect this constraint to capture
{\em precisely\/} complexity classes like $\C{NP}$ or $\C{NEXP}$.
That is, for any language $L$ that has an $\C{NP}$-parser, there exists
an R-AVG, say $G$, such that $L=L(G)$. Though our refinement, the
{\em honest parsability constraint\/} is probably not a property that
can be decided for arbitrary R-AVGs,
we show that R-AVGs can be equipped
with restricting mechanisms that enforce this property.
The techniques that prove Theorem~\ref{ravgl} and
Theorem~\ref{npishpavg} result from Johnson's work.
Therefore, the proofs of these theorems are deferred
to the appendices.

\section{Definitions and Notation}
\subsection{Attribute-Value Grammars}
The definitions in this section are in the spirit
of~\cite[Section~3.2]{johnson:88}
and~\cite[Sections~3--4]{smolka:92}.
Consider three sets of pairwise disjoint symbols.
\begin{list}{}{}
\item $A$, the finite set of constants, denoted ($a,b,c,\dots$)
\item $V$, the countable set of variables, denoted ($x,y,z,\ldots$)
\item $L$, the finite set of attributes, also called features,
denoted ($f,g,h,\ldots$)
\end{list}

\begin{define}
An {\em $f$-edge\/} from $x$ to $s$ is a triple $(x,f,s)$ such that $x$ is
a variable, $f$ is an attribute, and $s$ is a constant or a variable.
A {\em path\/}, $p$, is a, possibly empty, sequence of $f$-edges
$(x_1,f_1,x_2),(x_2,f_2,x_3),\ldots,(x_{n},f_n,s)$
in which the $x_i$ are
variables and $s$ is either a variable or a constant.
Often a path is denoted by the sequence of its edges' attributes,
in reversed order, e.g., $p = f_n\ldots f_1$.
Let $p$ be a path, $ps$ denotes the path that starts from $s$,
where $s$ is a constant only if $p$ is the empty path. If the path
is nonempty, $p = f_n\ldots f_1 \;(n geq 1)$, then $s$ is a variable.
For paths
$ps$ and $qt$ we write $ps\doteq qt$ iff $p$ and $q$ start in $s$
and $t$ respectively and end in the same variable or constant.
The expression $ps\doteq qs$ is called a {\em path equation\/}.
A {\em feature graph\/} is either a pair $(a,\emptyset)$,
or a pair $(x,E)$ where $x$ is the root and $E$ a finite set of
$f$-edges such that:
\begin{enumerate}
\item if $(y,f,s)$ and $(y,f,t)$ are in $E$, then $s=t$;
\item if $(y,f,s)$ is in $E$, then there
is a path from $x$ to $y$ in $E$.
\end{enumerate}
\end{define}

\begin{define}
An {\em attribute-value language\/} ${\cal A}(A,V,L)$ consists of
sets of logical formulas that describe feature graphs,
by expressing constraints on the type of paths that can exist within
the graphs.
\begin{itemize}
\item  The terms of an
attribute-value language ${\cal A}(A,V,L)$ are the
constants and the variables $s,t\in A\cup V$.
\item The formulas of an attribute-value language ${\cal A}(A,V,L)$
are path equations and Boolean combinations
of path equations. Thus all formulas are either $ps\doteq qt$, where
$ps$ and $qt$ are paths, or
$\phi \wedge \psi$, $\phi \vee \psi$, or $\neg \phi$, where $\phi$ and
$\psi$ are formulas.
\end{itemize}
\end{define}

Assume a finite set $\Lex$ (of lexical forms) and a finite set $\Cat$
(of categories).
$\Lex$ will play the role of the set of terminals and $\Cat$ will play the
role of the set of nonterminals in the productions.

\begin{define} A {\em constituent structure tree\/} (CST) is a labeled
tree in which the internal nodes are labeled with elements of Cat and the
leaves are labeled with elements of Lex.
\end{define}

\begin{define} Let $T$ be a constituent structure tree and $F$ be
a set of formulas in an attribute-value language  ${\cal A}(A,V,L)$.
An {\em annotated constituent structure tree\/} is a triple
$\mathopen{<}T,F,h\mathclose{>}$, where $h$ is a function that maps
internal nodes in $T$ onto variables in $F$.
\end{define}

\begin{define} A {\em lexicon\/} is a finite subset of $\Lex\times
\Cat\times  {\cal A}(A,\{x_0\},L)$. A set of
{\em syntactic rules} is a finite subset of
$\bigcup_{i\geq 1} \Cat\times\Cat^i\times{\cal A}(A,\{x_0,\ldots,x_i\},L)$.
An {\em attribute-value grammar\/} is a triple
$\mathopen{<}\mathord{\rm lexicon},\mathord{\rm rules},\mathord{\rm start}
\mathclose{>}$, where lexicon is a lexicon, rules is a set of
syntactic rules and start is an element of $\Cat$.
\end{define}

\begin{define}\hspace*{\fill}\\
\vspace{-2\topsep}
\begin{enumerate}
\item {\bf \cite[p~.150]{balcazar.ea:88} }
A class ${\cal C}$ of sets is {\em recursively presentable} iff there is an
effective enumeration $M_1, M_2, \ldots$ of deterministic Turing
machines which halt on all their inputs, and such that ${\cal C} = \{
L(M_i) \mid i = 1, 2, \ldots \}$.
\item We say that a class of grammars ${\cal G}$ is {\em recursively
presentable} iff the class of sets $\{L(G) \mid G \in {\cal G} \}$
is recursively presentable.
\end{enumerate}
\end{define}

\subsection{Restricted Attribute-Value Grammars}
The only formulas that are allowed in the attribute-value language
of restricted attribute-value grammars (R-AVGs)
are path-equations and conjunctions
of path-equations (i.e.~disjunctions and negations are out). We
will denote the attribute-value language of an R-AVG by
${\cal A'}(A,V,L)$ to make the distinction clear.
The CST of an R-AVG is produced by a chain- and $\epsilon$-rule free
regular grammar. The CST of an R-AVG can be either a left-branching
or a right-branching tree, since the grammar contains
at most one nonterminal in each rule.
\begin{define}
The set of syntactic rules of a restricted attribute-value
grammar is a subset of $\bigcup_{i\geq 1, k\leq 1}
\Cat\times\Lex^i\times \Cat^k\times {\cal A'}(A,\{x_0,x_k\},L)$.
A {\em restricted attribute-value grammar\/} is a pair
$\mathopen{<}\mathord{\rm rules},\mathord{\rm start}
\mathclose{>}$, where rules is a set of syntactic rules and start is
an element of $\Cat$.
\end{define}

\begin{define}
An R-AVG $\mathopen{<}\mathord{\rm rules},\mathord{\rm start}
\mathclose{>}$ {\em generates\/} an annotated constituent structure
tree $\mathopen{<}T,F,h\mathclose{>}$ iff
\begin{enumerate}
\item the root node of $T$ is start, and
\item every internal node of $T$ is licensed by a syntactic rule, and
\item the set $F$ is consistent, i.e., describes a feature graph.
\end{enumerate}
Let $\phi[x/y]$ stand for the formula $\phi$ in which all
variable $y$ is substituted for variable $x$.
An internal node $v$ of an annotated constituent structure tree is
{\em licensed\/} by a syntactic rule $(c_0, l_1,\ldots,l_i,\phi)$
iff
\begin{enumerate}
\item the node $v$ is labeled with category $c_0$, $h(v) = n_0$, and
\item all daughters of $v$ are leaves, which are labeled with
	$l_1 \ldots l_i$, and
\item $\phi[x_0/n_0]$ is in the set $F$.
\end{enumerate}
An internal node $v$ of an annotated constituent structure tree is
{\em licensed\/} by a syntactic rule $(c_0, l_1,\ldots,l_i, c_1,\phi)$
iff
\begin{enumerate}
\item the node $v$ is labeled with category $c_0$, $h(v) = n_0$, and
\item one of $v$'s daughters is an internal node, $v_1$, which is
	labeled with category $c_1$, and $h(v_1) = n_1$, and
\item the daughters of $v$ that are leaves are labeled with
	$l_1 \ldots l_i$, and
\item $\phi[x_0/n_0, x_1/n_1]$ is in the set $F$.
\end{enumerate}
\end{define}

\section{Weak Generative Capacity}\label{WGC}
In~\cite{lp-95-01},
it is shown that the recognition problem for
R-AVGs is $\C{NP}$-complete. This seems to indicate that although
the mechanism for generating CSTs in R-AVGs is extremely simple, the
generative capacity of R-AVGs is different from the generative
capacity of e.g., context free languages (CFLs), which have a polynomial
time parsing algorithm~\cite{earley:70}.
Yet, a priori, there may exist CFLs that do not have an R-AVG.

\begin{thm}\label{ravgl}
Let $L$ be a context free language. There exists an
R-AVG $G$ such that $L=L(G)$.
\end{thm}
\begin{proof} If $L$ is a context free language, then there exists
a context free grammar $G'$ in Greibach normal form such that $L=L(G')$.
{}From this grammar $G'$, we can construct a pushdown store
$M$ that accepts exactly the words in $L(G')=L$. Such a pushdown store
$M$ is actually a finite state automaton $M'$ with a stack $S$.
The finite state automaton $M'$ may be simulated by a
chain- and $\epsilon$-rule free regular grammar.
Furthermore, we can construct an attribute-value language
${\cal A'}(A,V,L)$ that simulates the stack $S$.
Thus it should be clear that there exists an R-AVG $G$ that
produces word $w$ iff $w \in L(G')$.
Details of this construction are deferred to Appendix~\ref{Greib}.
\end{proof}

{}From this we can draw the conclusion that the class of context
free languages is indeed a proper subset of the class of R-AVG languages,
unless $\C{P}=\C{NP}$.

\begin{thm}\label{PNP}
Let ${\cal C}$ be a recursively presentable class of grammars such that:
\begin{enumerate}
\item $G\in{\cal C}$ can be decided in time polynomial in $\lng{G}$
\item $G\stackrel{*}{\Rightarrow}w$ can be decided in time
polynomial in $\lng{G}+\lng{w}$.
\end{enumerate}
If every R-AVG $G$ has a grammar
in ${\cal C}$ then $\C{P}=\C{NP}$. In fact,
for every language $L$ in
$\C{NP}$ there is an explicit deterministic polynomial time algorithm.
\end{thm}
\begin{proof} Let $L$ be a language in $\C{NP}$ and $w\in \{0,1\}^*$.
Trautwein~\cite{lp-95-01} provided
an R-AVG $G$ and a reduction that maps any formula $F$ onto a string
$w_F$ s.t. $G\stackrel{*}{\Rightarrow}w_F$ iff $F\in\mathord{\it SAT}$.
It was also shown that any R-AVG has a nondeterministic polynomial time,
hence deterministic exponential time, recognition algorithm.
Suppose every R-AVG $G$ has a grammar
in ${\cal C}$.  Then there exists a $G' \in {\cal C}$
with $L(G')=L(G)$. We can decide
in polynomial time whether $w_F\in L(G)$ for any $w_F$. So,
$\C{P}=\C{NP}$.

If every R-AVG $G$ has a grammar in ${\cal C}$, then the algorithm
for deciding ``$w\in L$?'' consists of: use Cook's reduction
to produce a formula $F$ that is satisfiable iff $w\in L$; use
Trautwein's reduction to produce $w_F$ and R-AVG $G$;
enumerate grammars in $\cal C$ for the first grammar $G'$
that has a description of length less than $\log\log\lng{w}$
for which $L(G)\cap\{0,1\}^{\leq\log\log{\lng{w}}}=L(G')
\cap\{0,1\}^{\leq\log\log{\lng{w}}}$
accept iff $w\in L(G')$.
This gives a polynomial time algorithm that erroneously accepts or
rejects $w$ for only a finite number of strings $w$. The theorem now
follows from the fact that both $\C{P}$ and $\C{NP}$ are closed under
finite variation.
\end{proof}

\begin{cor} If R-AVGs generate only context free languages then
$\C{P}=\C{NP}$.
\end{cor}

In fact it can be shown directly that R-AVGs also produce non-context
free languages.

\begin{thm} The context sensitive language $\{a^nb^nc^n\}$ is
generated by an R-AVG.
\end{thm}
\begin{proof}(Sketch) Typically, the R-AVG that generates the language
$\{a^nb^nc^n\}$ first generates an amount of $a$'s then an amount
of $b$'s and finally an amount of $c$'s.
Let us assume that the grammar generates $i$ $a$'s. During the
derivation, the feature graph can be used to store the amount
of $a$'s that is produced. Once the grammar starts to produce
$b$'s , the feature graph will force the grammar to generate exactly
$i$ $b$'s and next to generate exactly $i$ $c$'s as well.
\end{proof}

\section{The Honest Parsability Constraint and Consequences}\label{HPC}
According to Theorem~\ref{PNP}, it is unlikely that the languages
generated by R-AVGs can be limited to those languages with a
polynomial time recognition algorithm.
Trautwein~\cite{lp-95-01} showed that all R-AVGs
have nondeterministic polynomial time algorithms. Is it perhaps
the case that any language that has a nondeterministic polynomial
time recognition algorithm can be generated by an R-AVG. Does
there exist a tight relation between time bounded machines and R-AVGs
as e.g., between LBAs and CSLs? The answer is that the off-line
parsability constraint that forces the R-AVG to have no chain-
or $\epsilon$-rules
is just too restrictive to allow such a connection.
The following trick to alleviate this problem has been observed earlier
in complexity theory. The off-line
parsability constraint(OLP)~\cite{johnson:88} relates
the amount of ``work'' done by the grammar to produce a string
linearly to the number of terminal symbols produced. It is therefore
a sort of honesty constraint that is also demanded of functions
that are used in e.g., cryptography. There the deal is, for each
polynomial amount of work done to compute the function at least
one bit of output must be produced. In such a way, for polynomial
time computable functions one can guarantee that the inverse of
the function is computable in nondeterministic polynomial time.

As a more liberal constraint on R-AVGs we propose an analogous
variation on the OLP
\begin{define} A grammar $G$ satisfies the Honest Parsability
Constraint(HPC) iff there exists a polynomial $p$ s.t. for each $w$
in $L(G)$ there exists a derivation with at most $p(\lng{w})$
steps.
\end{define}

{}From Smolka's algorithm and Trautwein's observation it trivially
follows that any attribute-value grammar that satisfies the
HPC (HP-AVG) has an $\C{NP}$ recognition algorithm. The problem with the
HPC is of course that it is not a syntactic property of grammars.
The question whether a given AVG satisfies the HPC (or the OLP for
that matter) may well be undecidable.
Nonetheless, we can produce a set of rules that,
when added to an attribute-value grammar {\em enforces\/} the
HPC. The newly produced language is then a subset of the old
produced language with an $\C{NP}$ recognition algorithm.
Because of the fact that our addition may simulate any polynomial
restriction, we regain the full class of AVG's that satisfy the HPC.
In fact
\begin{thm}
The class, P-AVGL, of languages produced by the HP-AVGs
is recursively presentable.
\end{thm}

We will give
a detailed construction of such a set of rules in
Appendix~\ref{hpavg}. The existence of such a set of rules
and the work of Johnson now gives the following theorem.

\begin{thm}\label{npishpavg}
For any language $L$ that has an $\C{NP}$ recognition
algorithm, there exists a
restricted attribute-value grammar $G$ that respects
the HPC and such that $L=L(G)$.
\end{thm}
\begin{proof}(Sketch)
Let $M$ be the Turing machine that decides $w\in L$. Use
a variation of Johnson's construction of a Turing machine
to create an R-AVG
that can produce any string $w$ that is recognized by $M$. Add the
set of rules that guarantee that only strings that can be produced
with a polynomial number of rules can be produced by the grammar.
\end{proof}

\section{Veer out the HPC}
Instead of creating a counter of logarithmic size as we do in
Appendix~\ref{hpavg}, it is quite straightforward to construct
a counter of linear size (or exponential size if there is enough
time). In fact, for well-behaved functions, the construction of a
counter gives a method to enforce any desired time bound constraint on
the recognition problem for attribute-value grammars.
For instance, for nondeterministic exponential time we could define
the Linear Dishonest Parsability Constraint (LDP) (allowing a linear
exponential number of steps) which would give.

\begin{thm} The class of languages generated by R-AVGs obeying the
LDP condition is exactly $\C{NE}$.
\end{thm}

\section*{Acknowledgements}
We are indebted to E.~Aarts and W.C.~Rounds for their valuable
suggestions on an early presentation of this work.

\appendix
\section{Simulating a Context Free Grammar in GNF}\label{Greib}
A context free grammar (CFG) is a quadruple
$\langle N, \Sigma, P, S \rangle$, where $N$ is a set of nonterminals,
$\Sigma$ is a set of terminals, $P$ is a set of productions,
and $S \in N$ is the start nonterminal.
A CFG is in Greibach normalform (GNF) if, and only if, the
productions are of one of the following
forms, where $a\in\Sigma, A\in N, A_1\ldots A_n \in N\setminus\{S\}$
and
$\epsilon$ the empty string (c.f., \cite{hopcroft.ea:79},
\cite{sudkamp:88}):
\begin{eqnarray*}
	A &\rightarrow &a\,A_1\ldots A_n	\\
	A &\rightarrow &a	\\
	S &\rightarrow &\epsilon
\end{eqnarray*}

Given a GNF $G = \langle N, \Sigma, P, S \rangle$,
we can construct a restricted attribute-value grammar (R-AVG)
$G'$ that simulates grammar $G$. R-AVG $G'$ consists
of the same set of nonterminals and terminals as GNF $G$.
The productions of R-AVG $G'$ are described by
Table~\ref{RulesSim}.
The only two attributes of R-AVG $G'$ are \fn{top} and
\fn{rest}. R-AVG $G'$ contains $|N| + 1$ atomic values, one atomic
value for each nonterminal and the special atomic value \$.
The R-AVG $G'$ uses the feature graph to
encode a push-down stack, similar to the encoding of a list.
The stack will be used to store the
nonterminals that still have to be rewritten.

The three syntactic abbreviations below are used to clarify the simulation.
We use represent a stack by a Greek letter, or a string of symbols;
the top of the stack is the leftmost symbol of the string.
Let $x_0$ encode a stack $\gamma$, then the formulas in
the abbreviation $\fn{push}(A_0\ldots A_n)$ express that $x_1$ encodes
a stack $A_0\ldots A_n\gamma$. Likewise, the formulas in the
abbreviation $\fn{pop}(A)$ express that $x_0$ encodes a stack
$A\gamma$, and $X_1$ encodes the stack $\gamma$. The abbreviation
\fn{empty-stack} expresses that $x_0$ encodes an empty stack.
\begin{eqnarray*}
\fn{push}(A_0\ldots A_n)	&\mbox{stands for}
	 &\fn{top}(x_1) \doteq A_0 \wedge 	\\
	&&\fn{top rest}(x_1) \doteq A_1 \wedge 	\\
	&&{\centering \vdots}\\
	&&\fn{top rest}^n(x_1) \doteq A_n \wedge\\
	&&\fn{rest}^{n+1}(x_1) \doteq x_0	\\
\fn{pop}(A)	&\mbox{stands for}
	 &\fn{top}(x_0) \doteq A \wedge	\\
	&&\fn{rest}(x_0) \doteq x_1	\\
\fn{empty-stack}	&\mbox{stands for}	&x_0 \doteq \$
\end{eqnarray*}

\begin{table}[ht]
\centering
\fbox{
\begin{minipage}{12cm}
\footnotesize
\[\begin{array}{llll}
\mbox{Productions of GNF $G$} &&\mbox{Productions of R-AVG $G'$} \\
S \rightarrow a A_1\ldots A_n	&\leadsto &S \rightarrow a A_1 \\
	&&\fn{push}(A_2 \ldots A_n) \wedge \fn{empty-stack}\\
A \rightarrow a A_1\ldots A_n	&\leadsto &A \rightarrow a A_1 \\
	&&\fn{push}(A_2 \ldots A_n) 	&(A \neq S) \\
S \rightarrow a		&\leadsto &S \rightarrow a B
	&\forall B \in N \setminus \{S\} \\
	&&\fn{pop}(B) \wedge \fn{empty-stack}\\
S \rightarrow a		&\leadsto &S \rightarrow a \\
	&&\fn{empty-stack}\\
A \rightarrow a		&\leadsto &A \rightarrow a B
	&\forall B \in N \setminus \{S\} \\
	&&\fn{pop}(B) &(A \neq S) \\
A \rightarrow a		&\leadsto &A \rightarrow a \\
	&&\fn{empty-stack} \\
S \rightarrow \epsilon	&\multicolumn{2}{l}{\mbox{neglected}}
\end{array}\]
\end{minipage}
}
\caption{Simulating productions of GNF $G$ by R-AVG $G'$\label{RulesSim}}
\end{table}

We have to prove that GNF $G$
and its simulation by R-AVG $G'$
generate (almost) the same language. Obviously, R-AVG $G'$
cannot generate the empty string. However,
for all non-empty strings the following theorem holds.
\begin{thm}
Start nonterminal $S$ of GNF $G$
derives string $\alpha$ ($\alpha \in \Sigma^+$)
if, and only if,
start nonterminal $S$ of R-AVG $G'$
derives string $\alpha$ with the empty stack.
\end{thm}
\begin{proof}
There are two cases to consider. First, $S$ derives string $\alpha$
in one step. Second, $S$ derives string $\alpha$ in more than one
step. The lemma below is needed in the proof of the second case.

\begin{description}
\item[Case I]
   Let start nonterminal $S$ derive string $\alpha$ in one step.
   GNF $G$ contains a production
   $S \rightarrow \alpha$ iff R-AVG $G'$
   contains a production $S \rightarrow \alpha$ with the equation
   \fn{empty-stack}.
   So, $S$ derives $\alpha$ in a
   derivation of GNF $G$ iff $S$ derives $\alpha$
   with an empty stack in the derivation of R-AVG $G'$.
\item[Case II]
   Initial nonterminal $S$ of GNF $G$ derives
   string $\alpha=\beta\beta'$ in more than one step iff there is a
   left-most derivation $S \stackrel{*}{\Rightarrow} \beta A
   \Rightarrow
   \beta\beta'$. GNF $G$ contains production
   $A \rightarrow \beta'$ iff R-AVG $G'$ contains production
   $A \rightarrow \beta'$ with the equation
   \fn{empty-stack}. By the next lemma: $S \stackrel{*}{\Rightarrow}
   \beta A$ iff $S \stackrel{*}{\Rightarrow} \beta A$ with
   the empty stack.
   Hence $S$ derives $\alpha$ for GNF $G$ iff
   $S$ derives $\alpha$ with empty stack for R-AVG $G'$.
\end{description}
\vspace*{-2\topsep}
\end{proof}

\begin{lem}
Start nonterminal $S$ derives $\alpha A\gamma$ ($\alpha \in \Sigma^+,
A\gamma \in (N \setminus \{S\})^+$) in a left-most derivation of
GNF $G$ if, and only if,
nonterminal $S$ derives $\alpha A$ with stack $\gamma\$$
(\$ is the bottom-of-stack symbol)
in the derivation of R-AVG $G'$.
\end{lem}
\begin{proof}
The lemma is proven by induction on the length of the derivation.
\begin{description}
\item[Basis]
If $S$ derives $\alpha A \gamma$ in one step, then
GNF $G$ contains production $S \rightarrow \alpha A \gamma$ and
R-AVG $G'$ contains production $S \rightarrow \alpha A$
with stack $\gamma\$$.
If $S$ derives $\alpha A$ with stack $\gamma\$$ in one step, then
R-AVG $G'$ contains production $S \rightarrow \alpha A$
with stack $\gamma\$$ and
GNF $G$ contains production $S \rightarrow \alpha A \gamma$.
\item[Induction]
The induction hypotheses states that $S \stackrel{n}{\Rightarrow} \alpha A
\gamma$ in GNF $G$ iff $S \stackrel{n}{\Rightarrow} \alpha A$ with stack
$\gamma\$$ in R-AVG $G'$. Next, we distinguish three cases.
\begin{enumerate}
\item
GNF $G$ contains a production $A \rightarrow a A_1A_2\ldots A_n$.
Hence there is a left-most derivation $S \stackrel{n+1}{\Rightarrow}
\alpha a A_1A_2 \ldots A_n\gamma$.
GNF $G$ contains the production $A \rightarrow
a A_1A_2 \ldots A_n$ iff R-AVG $G'$
contains a production $A \rightarrow a A_1$ with equation
\fn{push}($A_2\ldots A_n$).
Since the induction hypotheses states that there is a derivation
$S \stackrel{n}{\Rightarrow} \alpha A$ with stack $\gamma\$$, there is a
derivation
$S \stackrel{n+1}{\Rightarrow} \alpha a A_1$ with stack $A_2 \ldots
A_n\gamma\$$.
\item
GNF $G$ contains a production $A \rightarrow a$ and
$\gamma = B'\gamma'$.
Hence there is a left-most derivation $S \stackrel{n+1}{\Rightarrow}
\alpha a B' \gamma'$.
GNF $G$ contains the production $A \rightarrow a$
iff R-AVG $G'$ contains
productions $A \rightarrow a B$ with equation \fn{pop}(B),
for all $B \in N \setminus \{S\}$.
Hence by the induction hypotheses, there is a
derivation
$S \stackrel{n+1}{\Rightarrow} \alpha a B'$ with stack $\gamma'\$$.
\item
GNF $G$ contains a production $A \rightarrow a$ and
$\gamma = \epsilon$.
Then there is a left-most derivation $S \stackrel{n+1}{\Rightarrow}
\alpha a$.
GNF $G$ contains the production $A \rightarrow a$
iff R-AVG $G'$ contains production
$A \rightarrow a$ with equation \fn{empty-stack}.
Hence by the induction hypotheses, there is a derivation
$S \stackrel{n+1}{\Rightarrow} \alpha a$ with stack $\$$.
\end{enumerate}
\end{description}
\vspace*{-2\topsep}
\end{proof}

Because every context free language is generated by some GNF $G$,
every context free language is generated by some R-AVG $G'$.

\section{Constructing an Honestly Parsable Attribute-Value
Grammar}\label{hpavg}
In this section we show how to add a binary counter to an
attribute-value grammar (AVG). This counter enforces the
Honest-Parsability Constraint (HPC) upon the AVG. To keep
this section legible we sometimes use the
attribute-value matrices (AVMs) as descriptions.
In Section~\ref{create}, we show how to create
a counter for the AVG.
In Section~\ref{avg2hpc} we show how to extend
the syntactic rules and the lexicon
of the AVG.

\subsection{Arithmetic by AVGs}
We start with a little bit of arithmetic.
\paragraph{Natural numbers.}
The AVMs below encode natural numbers in binary
notation.  The sequences of attributes \fn{0} and \fn{1} in these
AVMs
encode natural numbers, from least- to most-significant bit.
The attribute \fn{v} has value 1 (or 0) if, and only if,
it has a sister attribute \fn{1} (or \fn{0}).
\begin{enumerate}
\item The AVMs {\footnotesize $\avm{\fn{v} &0\\ \fn{0} &\fv{+}}$}
	and {\footnotesize $\avm{\fn{v} &1\\ \fn{1} &\fv{+}}$}
	encode the natural numbers zero and one.
\item The AVMs {\footnotesize $\avm{\fn{v} &0\\ \fn{0} &[F]}$}
	and {\footnotesize $\avm{\fn{v} &1\\ \fn{1} &[F]}$} encode
	natural numbers iff the AVM $[F]$ encodes a natural number.
\end{enumerate}

\paragraph{Syntactic rules that tests two numbers for equality.}
Assume a nonterminal $A$ with some AVM
{\footnotesize $\avm{\fn{n} &[F]\\ \fn{m} &[H]}$},
where $[F]$ and $[H]$ encode
natural number $x$ and $y$, respectively.
We present one syntactic rule that derives from this
nonterminal $A$ a nonterminal $B$ with AVM
{\footnotesize $\avm{\fn{n} &[F]\\ \fn{m} &[H]}$} if $x = y$.
\begin{table}[ht]
\centering
\fbox{
\footnotesize
$\avg{t}{A &\rightarrow\: B \\
  &\fn{n}(x_0) \doteq \fn{m}(x_0)	\\
  &\wedge\, x_0 \doteq x_1	} $
}
\caption{The rule to test two numbers for equality.}
\end{table}

Clearly, this simple test takes one step. A more sophisticated
test, which also tests for inequality, would compare $[F]$
and $[G]$ bit-by-bit.  Such a test would take
$O(\min(\log(x),\log(y)))$ $=O(\min(\lng{[F]},\lng{[H]}))$
derivation steps.

\paragraph{Syntactic rules that multiply by two.}
Assume a nonterminal $A$ with some AVM
$\avm{\fn{n} &[F]}$, where $[F]$ encodes natural number $x$.
We present one syntactic rule that derives from this
nonterminal $A$ a nonterminal $B$ with the AVM
$\avm{\fn{n} &[H]}$, where $[H]$ encodes natural number $2x$.

The number \fn{n} in $[H]$ equals two times \fn{n} in $[F]$
if, and only if, the least-significant bit of \fn{n} in $[H]$
is 0, and the remaining bits form the same sequence as the
number \fn{n} in $[F]$. Multiplication by two takes one derivation
step.
\begin{table}[ht]
\centering
\fbox{
\footnotesize
$\avg{t}{A &\rightarrow\: B \\
  &\fn{v n}(x_1) \doteq 0	\\
  &\wedge\,\fn{n}(x_0) \doteq \fn{0 n}(x_1) }$
}
\caption{The rule to multiply by two.}
\end{table}

\paragraph{Syntactic rules that increments by one.}
Assume a nonterminal $A$ with some AVM
$\avm{\fn{n} &[F]}$, where $[F]$ encodes natural number $x$.
We present five syntactic rules that derive from this
nonterminal $A$ a nonterminal $C$ with AVM
$\avm{\fn{n} &[H]}$, where $[H]$ encodes natural number $x+1$.

The increment of \fn{n} requires two additional pointers in the
AVM of $A$: attribute \fn{p} points to the next bit
that has to be incremented; attribute \fn{q} points to the
most-significant bit of the (intermediate) result.
These additional pointers are hidden from the
AVMs of the nonterminals $A$ and $C$.

The five rules from Table~\ref{Tinc} increment \fn{n} by one.
Nonterminal $A$ rewrites, in one or more steps, to nonterminal
$C$, potentially through a number of nonterminals $B$.
\begin{table}[ht]
\centering
\fbox{
\begin{minipage}{11cm}
\footnotesize
\[\begin{array}{l|l|l}
\avg{t}{A' &\rightarrow\: C'\\
  &\fn{v n}(x_0) \doteq 0	\\
  &\wedge\,\fn{0 n}(x_0) \doteq \fn{1 n}(x_1) 	\\
  &\wedge\,\fn{v n}(x_1) \doteq 1	}
&\avg{t}{A' &\rightarrow\: B\\
  &\fn{v n}(x_0) \doteq 1	\\
  &\wedge\,\fn{1 n}(x_0) \doteq \fn{p}(x_1) 	\\
  &\wedge\,\fn{0 n}(x_1) \doteq \fn{q}(x_1) 	\\
  &\wedge\,\fn{v n}(x_1) \doteq 0 }
&\avg{t}{B &\rightarrow\: B\\
  &\fn{v p}(x_0) \doteq 1	\\
  &\wedge\,\fn{1 p}(x_0) \doteq \fn{p}(x_1) 	\\
  &\wedge\,\fn{n}(x_0) \doteq \fn{n}(x_1) 	\\
  &\wedge\,\fn{v q}(x_0) \doteq 0 	\\
  &\wedge\,\fn{0 q}(x_0) \doteq \fn{q}(x_1) }
\\ \hline
\avg{t}{B &\rightarrow\: C' \\
  &\fn{v p}(x_0) \doteq 0	\\
  &\wedge\,\fn{v q}(x_0) \doteq 1	\\
  &\wedge\,\fn{0 p}(x_0) \doteq \fn{1 q}(x_0) \\
  &\wedge\,\fn{n}(x_0) \doteq \fn{n}(x_1) 	}
&\avg{t}{B &\rightarrow\: C'	\\
  &\fn{v p}(x_0) \doteq 1	\\
  &\wedge\,\fn{1 p}(x_0) \doteq +	\\
  &\wedge\,\fn{n}(x_0) \doteq \fn{n}(x_1) 	\\
  &\wedge\,\fn{v q}(x_0) \doteq 0		\\
  &\wedge\,\fn{v 0 q}(x_0) \doteq 1	\\
  &\wedge\,\fn{1 0 q}(x_0) \doteq +	}
\end{array}\]
\end{minipage}
}
\caption{Five rules to increment \fn{n} by one.\label{Tinc}}
\end{table}

The first and fourth rule of Table~\ref{Tinc}
state that adding one to a zero bit sets
this bit to one and ends the increment. The second and third rule state
that adding one to a one bit sets this bit to zero and the increment
continues.  The fifth rule states that adding one to the
most-significant bit
sets this bit to zero and yields a new most-significant one bit.
We claim that $A \stackrel{*}{\Rightarrow} C$
takes $O(\log(x)) = O(\lng{[F]})$ derivation steps.

Rules, similar to the ones above, can be given that decrement
the attribute \fn{n} by one. We only have to take a little
extra care that the number 0 cannot be decremented.

\paragraph{Syntactic rules that sum two numbers.}
In this section we use the previous test and increment rules
(indicated by =).
Assume a nonterminal $A$ with some AVM
{\footnotesize $\avm{\fn{n} &[F]\\ \fn{m} &[H]}$},
where $[F]$ and $[H]$ encode
natural number $x$ and $y$, respectively.
We present syntactic rules (Table~\ref{Tcovsum}--\ref{Tstopsum})
that derive from this
nonterminal $A$ a nonterminal $C$ with AVM
{\footnotesize $\avm{\fn{n} &[F']\\ \fn{m} &[H]}$},
 where $[F']$ encodes the natural number $x + y$.

\begin{table}[htb]
\centering
\fbox{
\begin{minipage}{7cm}
\footnotesize
\[\begin{array}{l|l}
\avg{t}{A &\rightarrow\: A'\\
  &\fn{m}(x_0) \doteq \fn{m}(x_1) \\
  &\wedge\,\fn{n}(x_0) \doteq \fn{p}(x_1) \\
  &\wedge\,\fn{m}(x_1) \doteq \fn{q}(x_1) \\
  &\wedge\,\fn{r}(x_1) \doteq \fn{n}(x_1) }
&\avg{t}{C' &\rightarrow\: C\\
  &\fn{n}(x_0) \doteq \fn{n}(x_1) \\
  &\wedge\,\fn{m}(x_0) \doteq \fn{m}(x_1) }
\end{array}\]
\end{minipage}
}
\caption{Two rules to hide the auxiliary pointers.\label{Tcovsum}}
\end{table}

\begin{table}[htb]
\centering
\fbox{
\begin{minipage}{11cm}
\footnotesize
\[\begin{array}{l|l|l}
\avg{t}{A' &\rightarrow\: A'\\
  &\fn{v p}(x_0) \doteq 0 \\
  &\wedge\,\fn{v q}(x_0) \doteq 0 \\
  &\wedge\,\fn{v r}(x_0) \doteq 0 \\
  &\wedge\,\fn{0 p}(x_0) \doteq \fn{p}(x_1) \\
  &\wedge\,\fn{0 q}(x_0) \doteq \fn{q}(x_1) \\
  &\wedge\,\fn{0 r}(x_0) \doteq \fn{r}(x_1) \\
  &\wedge\,\fn{n}(x_0) \doteq \fn{n}(x_1) \\
  &\wedge\,\fn{m}(x_0) \doteq \fn{m}(x_1) }
&\avg{t}{A' &\rightarrow\: A'\\
  &\fn{v p}(x_0) \doteq 1 \\
  &\wedge\,\fn{v q}(x_0) \doteq 0 \\
  &\wedge\,\fn{v r}(x_0) \doteq 1 \\
  &\wedge\,\fn{1 p}(x_0) \doteq \fn{p}(x_1) \\
  &\wedge\,\fn{0 q}(x_0) \doteq \fn{q}(x_1) \\
  &\wedge\,\fn{1 r}(x_0) \doteq \fn{r}(x_1) \\
  &\wedge\,\fn{n}(x_0) \doteq \fn{n}(x_1) \\
  &\wedge\,\fn{m}(x_0) \doteq \fn{m}(x_1) }
&\avg{t}{A' &\rightarrow\: A'\\
  &\fn{v p}(x_0) \doteq 0 \\
  &\wedge\,\fn{v q}(x_0) \doteq 1 \\
  &\wedge\,\fn{v r}(x_0) \doteq 1 \\
  &\wedge\,\fn{0 p}(x_0) \doteq \fn{p}(x_1) \\
  &\wedge\,\fn{1 q}(x_0) \doteq \fn{q}(x_1) \\
  &\wedge\,\fn{1 r}(x_0) \doteq \fn{r}(x_1) \\
  &\wedge\,\fn{n}(x_0) \doteq \fn{n}(x_1) \\
  &\wedge\,\fn{m}(x_0) \doteq \fn{m}(x_1) }
\\ \hline
\avg{t}{B &\rightarrow\: B\\
  &\fn{v p}(x_0) \doteq 1 \\
  &\wedge\,\fn{v q}(x_0) \doteq 0 \\
  &\wedge\,\fn{v r}(x_0) \doteq 0 \\
  &\wedge\,\fn{1 p}(x_0) \doteq \fn{p}(x_1) \\
  &\wedge\,\fn{0 q}(x_0) \doteq \fn{q}(x_1) \\
  &\wedge\,\fn{0 r}(x_0) \doteq \fn{r}(x_1) \\
  &\wedge\,\fn{n}(x_0) \doteq \fn{n}(x_1) \\
  &\wedge\,\fn{m}(x_0) \doteq \fn{m}(x_1) }
&\avg{t}{B &\rightarrow\: B\\
  &\fn{v p}(x_0) \doteq 0 \\
  &\wedge\,\fn{v q}(x_0) \doteq 1 \\
  &\wedge\,\fn{v r}(x_0) \doteq 0 \\
  &\wedge\,\fn{0 p}(x_0) \doteq \fn{p}(x_1) \\
  &\wedge\,\fn{1 q}(x_0) \doteq \fn{q}(x_1) \\
  &\wedge\,\fn{0 r}(x_0) \doteq \fn{r}(x_1) \\
  &\wedge\,\fn{n}(x_0) \doteq \fn{n}(x_1) \\
  &\wedge\,\fn{m}(x_0) \doteq \fn{m}(x_1) }
&\avg{t}{B &\rightarrow\: B\\
  &\fn{v p}(x_0) \doteq 1 \\
  &\wedge\,\fn{v q}(x_0) \doteq 1 \\
  &\wedge\,\fn{v r}(x_0) \doteq 1 \\
  &\wedge\,\fn{1 p}(x_0) \doteq \fn{p}(x_1) \\
  &\wedge\,\fn{1 q}(x_0) \doteq \fn{q}(x_1) \\
  &\wedge\,\fn{1 r}(x_0) \doteq \fn{r}(x_1) \\
  &\wedge\,\fn{n}(x_0) \doteq \fn{n}(x_1) \\
  &\wedge\,\fn{m}(x_0) \doteq \fn{m}(x_1) }
\end{array}\]
\end{minipage}
}
\caption{Rules when the carry bit is not changed.}
\end{table}

\begin{table}[pt]
\centering
\fbox{
\begin{minipage}{7cm}
\footnotesize
\[\begin{array}{l|l}
\avg{t}{A' &\rightarrow\: B\\
  &\fn{v p}(x_0) \doteq 1 \\
  &\wedge\,\fn{v q}(x_0) \doteq 1 \\
  &\wedge\,\fn{v r}(x_0) \doteq 0 \\
  &\wedge\,\fn{1 p}(x_0) \doteq \fn{p}(x_1) \\
  &\wedge\,\fn{1 q}(x_0) \doteq \fn{q}(x_1) \\
  &\wedge\,\fn{0 r}(x_0) \doteq \fn{r}(x_1) \\
  &\wedge\,\fn{n}(x_0) \doteq \fn{n}(x_1) \\
  &\wedge\,\fn{m}(x_0) \doteq \fn{m}(x_1) }
&\avg{t}{B &\rightarrow\: A'\\
  &\fn{v p}(x_0) \doteq 0 \\
  &\wedge\,\fn{v q}(x_0) \doteq 0 \\
  &\wedge\,\fn{v r}(x_0) \doteq 1 \\
  &\wedge\,\fn{0 p}(x_0) \doteq \fn{p}(x_1) \\
  &\wedge\,\fn{0 q}(x_0) \doteq \fn{q}(x_1) \\
  &\wedge\,\fn{1 r}(x_0) \doteq \fn{r}(x_1) \\
  &\wedge\,\fn{n}(x_0) \doteq \fn{n}(x_1) \\
  &\wedge\,\fn{m}(x_0) \doteq \fn{m}(x_1) }
\end{array}\]
\end{minipage}
}
\caption{Rules when the carry bit is changed.}
\end{table}

\begin{table}[htb]
\centering
\fbox{
\begin{minipage}{10cm}
\footnotesize
\[\begin{array}{l|l|l}
\avg{t}{A' &\rightarrow\: C'\\
  &\fn{p}(x_0) \doteq + \\
  &\wedge\,\fn{q}(x_0) = i \\
  &\wedge\,\fn{r}(x_0) = j \\
  &\wedge\,i = j \\
  &\wedge\,x_0 \doteq x_1 }
&\avg{t}{A' &\rightarrow\: C'\\
  &\fn{p}(x_0) = i \\
  &\wedge\,\fn{q}(x_0) \doteq + \\
  &\wedge\,\fn{r}(x_0) = j \\
  &\wedge\,i = j \\
  &\wedge\,x_0 \doteq x_1 }
&\avg{t}{A' &\rightarrow\: C'\\
  &\fn{p}(x_0) \doteq + \\
  &\wedge\,\fn{q}(x_0) \doteq + \\
  &\wedge\,\fn{r}(x_0) \doteq + \\
  &\wedge\,x_0 \doteq x_1 }
\\ \hline
\avg{t}{B &\rightarrow\: C'\\
  &\fn{p}(x_0) \doteq + \\
  &\wedge\,\fn{q}(x_0) = z \\
  &\wedge\,\fn{r}(x_0) = z+1 \\
  &\wedge\,x_0 \doteq x_1 }
&\avg{t}{B &\rightarrow\: C'\\
  &\fn{p}(x_0) = z \\
  &\wedge\,\fn{q}(x_0) \doteq + \\
  &\wedge\,\fn{r}(x_0) = z+1 \\
  &\wedge\,x_0 \doteq x_1 }
&\avg{t}{B &\rightarrow\: C'\\
  &\fn{p}(x_0) \doteq + \\
  &\wedge\,\fn{q}(x_0) \doteq + \\
  &\wedge\,\fn{v r}(x_0) \doteq 1 \\
  &\wedge\,\fn{1 r}(x_0) \doteq + \\
  &\wedge\,x_0 \doteq x_1 }
\end{array}\]
\end{minipage}
}
\caption{Rules that stop the summation.\label{Tstopsum}}
\end{table}

The increment of \fn{n} by \fn{m} is similar to the
increment by one. Here, three additional pointers are required:
the attributes \fn{p} and \fn{q} point to the bits in \fn{n} and
\fn{m} respectively that have to be summed next; attribute \fn{r}
points to the most-significant bit of the (intermediate) result.
In the addition two states
are distinguished. In the one state, the carry bit is zero, indicated
by nonterminal $A'$. In the other state, the carry bit is one,
indicated by nonterminal $B$.
We claim that $A \stackrel{*}{\Rightarrow} C$
takes $O(\max(\log(x), \log(y))) = O(\max(\lng{[F]},\lng{[H]}))$
derivation steps.

\paragraph{Syntactic rules that sum a sequence of numbers.}
In this section we use the previous summation rules
(indicated by =).
Assume a nonterminal $A$ with some AVM
$\avm{\fn{l} &[F']}$,
where $[F']$ encodes a list of numbers. To wit
\[ [F'] \;=\;
\mbox{\footnotesize
	$\avm{\fn{f} &[G_1] \\ \fn{r} &\avm{\fn{f} &[G_2] \\
	\fn{r} &\ldots\avm{\fn{f} &[G_n] \\ \fn{r} &\fv{+}}}}$ } \]
where $[G_i]$ encodes natural number $x_i$.
We present syntactic rules (Table~\ref{Tlist}) that derive from this
nonterminal $A$ a nonterminal $B$ with AVM
{\footnotesize $\avm{\fn{suml} &[F]\\ \fn{l} &[F']}$},
where $[F]$ encodes the natural number $\Sigma_i x_i$.

The summation requires an additional pointer in the
AVM $[F']$: attribute \fn{p} points to the next element
in the list that has to be summed.
We claim that $A \stackrel{*}{\Rightarrow} B$
takes $O(\Sigma_i\,\log(x_i)) = O(\lng{[F']})$ derivation steps.
\begin{table}[ht]
\centering
\fbox{
\begin{minipage}{11cm}
\footnotesize
\[\begin{array}{l|l|l}
\avg{t}{A &\rightarrow\: A'\\
  &\fn{v n}(x_1) \doteq 0 \\
  &\wedge\,\fn{0 n}(x_1) \doteq + \\
  &\wedge\,\fn{l}(x_0) \doteq \fn{l}(x_1) \\
  &\wedge\,\fn{l}(x_0) \doteq \fn{p}(x_1) }
&\avg{t}{A' &\rightarrow\: A'\\
  &\fn{suml}(x_0) = y \\
  &\wedge\,\fn{f p}(x_0) = z \\
  &\wedge\,\fn{suml}(x_1) = y+z \\
  &\wedge\,\fn{r p}(x_0) \doteq \fn{p}(x_1) \\
  &\wedge\,\fn{l}(x_0) \doteq \fn{l}(x_1) }
&\avg{t}{A' &\rightarrow\: B\\
  &\fn{p}(x_0) \doteq + \\
  &\wedge\,\fn{suml}(x_0) \doteq \fn{suml}(x_1) \\
  &\wedge\,\fn{l}(x_0) \doteq \fn{l}(x_1) }
\end{array}\]
\end{minipage}
}
\caption{Three rules that sum a list of numbers.\label{Tlist}}
\end{table}

\subsection{Creating a counter of logarithmic size\label{create}}
Create an AVM of the following form:
{\footnotesize \[\avm{\fn{counter} &\avm{
	\fn{size} &\avm{\fn{1}\cup\fn{0} &\ldots[\fn{1}\;+]} \\
	\fn{n} &\avm{\fn{v} &1 \cup 0\\ \fn{1}\cup\fn{0}
		&\avm{\fn{v} &1 \cup 0\\ \ldots &[\fn{1}\;+]}} \\
	\fn{m} &\avm{\fn{v} &1 \cup 0\\ \fn{1}\cup\fn{0}
		&\avm{\fn{v} &1 \cup 0\\ \ldots &[\fn{1}\;+]}} \\
	\fn{poly} &\avm{\fn{1}\cup\fn{0} &\ldots[\fn{1}\;+]}} } \] }

Attribute \fn{counter} is used to distinguish the AVMs
that encodes the counter from those in the
original attribute-value grammar. We will neglect the
attribute \fn{counter} in the remainder of this section, because it is
not essential here.
The attributes \fn{size}, \fn{n}, \fn{m} and \fn{poly}
encode natural numbers.
The attribute \fn{size} records the size of the string that will be
generated.  The attribute \fn{poly} records the maximum number of
derivation steps that is allowed for a string of size \fn{size}.
The attributes \fn{n} and \fn{m} are auxiliary numbers.

The construction of the counter starts with an initiation-step.
The further construction of the counter consists of cycles of two
phases. Each cycle starts in nonterminal $A$.

\paragraph{Initiation step and first phase.}
The initiation-step sets the numbers \fn{size} and \fn{n} to 0,
and the numbers \fn{m} and \fn{poly} to 1.
In the first phase of each cycle, the numbers \fn{size} and \fn{n}
are incremented by 1.
\begin{table}[ht]
\centering
\fbox{
\begin{minipage}{9cm}
\footnotesize
\[\begin{array}{l|l}
\avg{t}{S &\rightarrow\: A\\
  &\fn{v size}(x_1)\doteq 0\\
  &\wedge\,\fn{0 size}(x_1)\doteq + \\
  &\wedge\,\fn{v n}(x_1)\doteq 0\\
  &\wedge\,\fn{0 n}(x_1)\doteq + \\
  &\wedge\,\fn{v m}(x_1)\doteq 1 \\
  &\wedge\,\fn{1 m}(x_1)\doteq + \\
  &\wedge\,\fn{1 poly}(x_1)\doteq + }
&\avg{t}{A &\rightarrow\: B	\\
  &\fn{size}(x_0) = x \\
  &\wedge\,\fn{size}(x_1) = x+1 \\
  &\wedge\,\fn{n}(x_0) = y \\
  &\wedge\,\fn{n}(x_1) = y+1 \\
  &\wedge\,\fn{m}(x_0)\doteq \fn{m}(x_1) \\
  &\wedge\,\fn{poly}(x_0)\doteq \fn{poly}(x_1) }
\end{array}\]
\end{minipage}
}
\caption{Initiation-step and first phase.\label{T1st}}
\end{table}

\paragraph{The second phase of the cycle.}
In this phase the numbers \fn{n} and \fn{m} are compared.
If \fn{n} is twice \fn{m}, then ($i$) number
\fn{poly} is extended by $k$ bits, ($ii$) number \fn{m} is doubled,
and ($iii$) number \fn{n} is set to 0.
If \fn{n} is less than twice \fn{m}, nothing happens.

The left rule of the second phase doubles the number \fn{m} in
the second and the third equation.
The test ``Is \fn{n} equal to 2\fn{m}?'' therefore reduces to one
(the first) equation.
The fourth equation extend the number \fn{poly} with $k$ bits.
The fifth and sixth equations set the number \fn{n} to 0.

The right rule is always applicable. If the right rule is used where
the left rule was applicable, then the number \fn{n} will never be
equal to $2\fn{m}$ in the rest of the derivation. Thus \fn{poly} will
not be extended any more.

\begin{table}[ht]
\centering
\fbox{
\begin{minipage}{7cm}
\footnotesize
\[\begin{array}{l|l}
\avg{t}{B &\rightarrow\: A \\
  &\fn{n}(x_0) = \fn{m}(x_1)	\\
  &\wedge\fn{m}(x_0) = x \\
  &\wedge\fn{m}(x_1) = 2x	\\
  &\wedge\fn{poly}(x_0) \doteq \fn{0$^k$ poly}(x_1)	\\
  &\wedge\fn{v n}(x_1) \doteq 0			\\
  &\wedge\fn{0 n}(x_1) \doteq +	}
&\avg{t}{B &\rightarrow\: A \\
  &x_0 \doteq x_1	}
\end{array}\]
\end{minipage}
}
\caption{The second phase.\label{T2nd}}
\end{table}

We claim that the left rule appears $\log(n)$ times and the right
rule $O(n)$ times in a derivation for input of size $n$.
Obviously, the number \fn{poly} is $O(2^{k\log{i}}) = O(i^k)$ when
the number \fn{size} is $i$.

\subsection{From AVG to HP-AVG\label{avg2hpc}}
In this section we show how to transform an AVG
into an AVG that satisfies the HPC (HP-AVG).
Since all computation steps of the HP-AVG
only require a linear amount of derivation steps,
total derivations of HP-AVGs have polynomial length.

We can divide the attributes of the HP-AVG into two groups. The
attributes that encode the counters, and the attributes of the
original AVG. The former will be embedded under the attribute
\fn{counter}, the latter under the attribute \fn{grammar}.
In the sequel, we mean by $\phi|\fn{grammar}$ the formula $\phi$
embedded under the attribute \fn{grammar}, i.e., the formula
obtained from $\phi$ by substituting the variables $x_i$ by
$\fn{grammar}(x_i)$.

The HP-AVG is obtained from the AVG in three steps: change
the start nonterminal, the lexicon and the syntactic rules.
First, the HP-AVG contains the rules of the previous section, which
construct the counter. The nonterminal $S$ from Table~\ref{T1st} is the
start nonterminal of the HP-AVG. For the nonterminal $A$ the start
nonterminal of the AVG is taken. Nonterminal $B$ from Table~\ref{T2nd}
is a fresh nonterminal, not occurring in the AVG.

Second, the HP-AVG contains an extension of the lexicon of the AVG.
The entries of the lexicon are extended in the following way. The size
of the lexical form is set to one, and the amount of derivation steps
is zero. Thus, if $(w,X,\phi)$ is the lexicon of the AVG,
then $(w,X,\psi)$ is the lexicon of the HP-AVG, where
\begin{eqnarray*}
\psi &= &\phi|\fn{grammar}	\\
&\wedge	&\fn{v size counter}(x_0) \doteq 1 \\
&\wedge	&\fn{1 size counter}(x_0) \doteq + \\
&\wedge	&\fn{poly counter}(x_0) \doteq +
\end{eqnarray*}

Third, the HP-AVG contains extensions of the syntactic rules of the AVG.
The syntactic rules are extended in the following way. The numbers
\fn{poly} and \fn{size} of the daughter nonterminals are collected in
the lists \fn{plist} and \fn{slist}. Both lists are summed.
The number \fn{size} of the mother nonterminal is equal to
the sum of \fn{size}'s,
and the number \fn{poly} of the mother nonterminal is one
more than the sum of \fn{poly}'s.
Thus, if $(X_0,X_1,\ldots,X_n, \phi)$
is a syntactic rule of the AVG, then $(X_0,X_1,\ldots,X_n, \psi)$
is a syntactic rule of the HP-AVG, where
\begin{eqnarray*}
\psi &= &\phi|\fn{grammar}	\\
&\wedge &\fn{sums counter}(x_0) = \Sigma\,\fn{slist counter}(x_0)  \\
&\wedge &\fn{size counter}(x_0) = \fn{sums counter}(x_0) \\
&\wedge &\fn{sump counter}(x_0) = \Sigma\,\fn{plist counter}(x_0)  \\
&\wedge &\fn{sump counter}(x_0) = y \\
&\wedge &\fn{poly counter}(x_0) = y+1 \\
&\wedge	&\fn{f r$^i$ slist counter}(x_0) \doteq \fn{size counter}(x_i)
\;\; (0 \leq i < n)	\\
&\wedge	&\fn{r$^n$ slist counter}(x_0) \doteq + \\
&\wedge	&\fn{f r$^i$ plist counter}(x_0) \doteq \fn{poly counter}(x_i)
\;\; (0 \leq i < n)	\\
&\wedge	&\fn{r$^n$ plist counter}(x_0) \doteq +
\end{eqnarray*}

Now, a derivation for the HP-AVG starts with a nondeterministic
construction of a counter \fn{size} with value $n$ and a counter
\fn{poly} with value $O(n^k)$. Then, the derivation of the original
AVG is simulated, such that
($i$)
 the mother nonterminal produces a string of size $n$ if, and only if
 the daughter nonterminals together produce a string of size $n$, and
($ii$)
 the mother nonterminal makes $n^k+1$ derivation steps if, and only if
 the daughter nonterminals together make $n^k$ derivation steps.

\end{document}